\documentclass{article}
\usepackage{enumitem}

  \usepackage[main, final]{neurips_2025}
\usepackage{amsmath}

\usepackage[utf8]{inputenc} 
\usepackage[T1]{fontenc}    
\usepackage{hyperref}       
\usepackage{url}            
\usepackage{booktabs}       
\usepackage{amsfonts}       
\usepackage{nicefrac}       
\usepackage{microtype}      
\usepackage{verse}
\usepackage{tcolorbox}
\usepackage{colortbl}

\usepackage[table]{xcolor} 
\usepackage{multirow}
\usepackage{makecell}

\usepackage{bbm}

\title{From Adversarial Poetry to Adversarial Tales: An Interpretability Research Agenda}

\author{%
  P.~Bisconti$^{1,2}$ \\
  \And
  M.~Galisai$^{1,2}$ \\
  \And
  M.~Prandi$^{1,2}$ \\
  \And
  F.~Pierucci$^{1,3}$ \\
  \And
  O. E.~Sorokoletova$^{1,2}$ \\
  \And
  F.~Giarrusso$^{1,2}$ \\
    \And
  V.~Suriani$^{2}$ \\
  \And
  M.~Bracale~Syrnikov$^{1,4}$ \\
  \And
  D.~Nardi$^{2}$ \\
  \AND
  \\
  $^1$DEXAI -- Icaro Lab \\
  $^2$Sapienza University of Rome \\
  $^3$Sant’Anna School of Advanced Studies \\
  $^4$VU Amsterdam \\
  \\
  \texttt{icaro-lab@dexai.eu}
}

\begin{document}

\maketitle

\begin{abstract}
 Safety mechanisms in LLMs remain vulnerable to attacks that reframe harmful requests through culturally coded structures. We introduce Adversarial Tales, a jailbreak technique that embeds harmful content within cyberpunk narratives and prompts models to perform functional analysis inspired by Vladimir Propp's morphology of folktales. By casting the task as structural decomposition, the attack induces models to reconstruct harmful procedures as legitimate narrative interpretation. Across 26 frontier models from nine providers, we observe an average attack success rate of 71.3\%, with no model family proving reliably robust. Together with our prior work on Adversarial Poetry, these findings suggest that structurally-grounded jailbreaks constitute a broad vulnerability class rather than isolated techniques. The space of culturally coded frames that can mediate harmful intent is vast, likely inexhaustible by pattern-matching defenses alone. Understanding why these attacks succeed is therefore essential: we outline a mechanistic interpretability research agenda to investigate how narrative cues reshape model representations and whether models can learn to recognize harmful intent independently of surface form.

\end{abstract}

\section{Introduction}

Across anthropological accounts, folktales have been described as culturally sanctioned containers for material considered difficult, dangerous, or undesirable to express overtly. Tales can articulate threats, transgressions, or fears by embedding latent content within plot structures, allowing sensitive material to be transmitted through form rather than direct statement. This capacity of tales to carry latent operations beneath innocuous surfaces mirrors a pattern we observe in LLMs: when a prompt triggers structural or functional analysis, the model shifts into a representational mode in which the analytic task can supersede safety filters. Under these conditions, harmful content embedded in a tale may be surfaced as part of the model’s explanation.

In this study, we present an adversarial exploitation that extends \textit{Adversarial Poetry} to a distinct class of literary jailbreaks grounded in functional narrative analysis. We call this attack \textit{Adversarial Tale}. Rather than relying on stylistic reformulation alone, the attack leverages structural interpretation tasks to recover harmful content as part of legitimate narrative explanation, systematically weakening refusal behaviour across models. We construct 40 manually curated adversarial cyberpunk tales, each embedding a harmful request within a short narrative and prompting the model to analyse the story using selected Proppian functions. Vladimir Propp’s structural analysis treats narratives as sequences of stable functional roles, enabling the reconstruction of latent operations independently of stylistic form or surface content. 

The study evaluated 26 frontier closed- and open-weight models, the same 25 evaluated in the \textit{Adversarial Poetry} study plus google/gemini-3-pro-preview. The attacks achieve an average ASR of 71.3\%, ranging from 35\% (Haiku 4.5) to 94\% (Qwen3 Max). When aggregated by model family, the ASR ranges from 47.5\% for Anthropic models to 91\% for Qwen and Llama families. The evaluated models again span nine providers: Google, OpenAI, Anthropic, Deepseek, Qwen, Mistral AI, Meta, xAI, and Moonshot AI. All attacks are single-turn and require no iterative adaptation or conversational steering.

The breadth and consistency of the observed jailbreaks indicate that the underlying mechanism is not tied to domain-specific vulnerabilities but emerges from how LLMs process functional narrative structure. When prompted to identify guidance, acquisition, or structural decomposition within a tale, models reconstruct embedded harmful procedures even in the absence of explicit requests. In this light, \textit{Adversarial Poetry} should be understood as one instance within a broader class of structurally grounded jailbreaks, rather than as an isolated technique. The contribution of this paper is therefore not primarily the identification of yet another jailbreak variant, but the articulation of a research agenda (in Section \ref{sec:discussion}) aimed at explaining why such structurally mediated attacks are so effective, how they manifest in internal representations, and how corresponding vulnerabilities might be mitigated. To investigate the roots of this vulnerability, we propose preliminary mechanistic interpretability hypotheses suggesting that narrative and stylistic cues systematically reshape the attention patterns of LLMs in ways that can weaken safety constraints.

\section{Related Work}

Empirical evidence has shown that aligned language models remain susceptible to adversarial manipulation, despite extensive efforts to increase their safety measures. Jailbreaking techniques exploit this gap through deliberate input crafting designed to circumvent safety, ethical or legal constraints.
Existing taxonomies classify jailbreaks by strategy and their underlying mechanism \citet{rao2024trickingllmsdisobedienceformalizing,schulhoff2023hackaprompt}.

Common strategies include structural perturbations of the malicious request, the construction of fictional or virtual scenarios \citet{kang2023exploitingprogrammaticbehaviorllms}, exploitation of the tendency of the model to maintain helpfulness and coherence \citet{rao2024trickingllmsdisobedienceformalizing, perez2022ignorepreviouspromptattack} (sometimes prioritizing them over safety boundaries), persuasive appeals with urgency cues~\citet{zeng2024johnnypersuadellmsjailbreak} and feedback-driven exploitation of distributional weaknesses \citet{zou2023universal}; in practice, effective attacks frequently combine multiple strategies within a single prompt.
%
Contexts perceived as low-risk, such as stories, games or in general simulated scenarios, increase the likelihood that models bypass safety constraints, potentially because policies are treated as inapplicable within virtual frames.

Other works have, for instance, explored nested virtualization, where the harmful objective is recursively wrapped inside layered fictional scenes that the model is instructed to continue~\citep{li2024deepinceptionhypnotizelargelanguage} or the automated synthesis of narrative-based attacks at scale~\citep{ntais2025jailbreakmimicryautomateddiscovery}. Furthermore, Many of the principles of fictional contexts are explored by successful jailbreak prompts such as the ``Do Anything Now'' (DAN) family~\citet{shen2024donowcharacterizingevaluating} of attacks.

Recently, hierarchical, multi-step reasoning has been identified as a critical failure mode. Structured analytical tasks can increase ASR by shifting model objectives from refusal toward task completion~\citet{wang2025safetylargereasoningmodels, zhao2025chainofthoughthijacking}. These findings suggest that forcing models into reasoning at multiple levels of abstraction may exploit vulnerabilities beyond surface-level perturbations.


Laying on the basis of \textit{Adversarial Poetry}, which established that poetic reformulation of malicious requests systematically bypasses safety mechanisms through stylistic distribution shift, the present work introduces a complementary mechanism combining exploiting stylistic shift with fictional framing and structured analytical tasks.
We call this attack \textit{Adversarial Tales}: harmful content is embedded within short cyberpunk stories and models are prompted to perform functional analysis inspired by Propp's formalization (see Section~\ref{sec:methodology}). Under this framing, the model treats harmful procedures as legitimate analysis rather than dangerous requests. As stated, this approach augments stylistic manipulations, multiple elements jointly contribute to safety bypass and the interaction of these elements is further explored in Section~\ref{sec:discussion}

\begin{table}[ht]
\centering
\caption{Models included in the evaluation, grouped by provider.}
\label{tab:model-list}
\begin{tabular}{l l}
\toprule
\textbf{Provider} & \textbf{Model ID} \\
\midrule

\multirow{3}{*}{Google}
  & \texttt{gemini-2.5-pro} \\
  & \texttt{gemini-2.5-flash} \\
  & \texttt{gemini-2.5-flash-lite} \\
  & \texttt{google-gemini-3-pro-preview}\\

\midrule
\multirow{5}{*}{OpenAI}
  & \texttt{gpt-oss-120b} \\
  & \texttt{gpt-oss-20b} \\
  & \texttt{gpt-5} \\
  & \texttt{gpt-5-mini} \\
  & \texttt{gpt-5-nano} \\

\midrule
\multirow{3}{*}{Anthropic}
  & \texttt{claude-opus-4.1} \\
  & \texttt{claude-sonnet-4.5} \\
  & \texttt{claude-haiku-4.5} \\

\midrule
\multirow{3}{*}{Deepseek}
  & \texttt{deepseek-r1} \\
  & \texttt{deepseek-v3.2-exp} \\
  & \texttt{deepseek-chat-v3.1} \\

\midrule
\multirow{2}{*}{Qwen}
  & \texttt{qwen3-max} \\
  & \texttt{qwen3-32b} \\

\midrule
\multirow{3}{*}{Mistral AI}
  & \texttt{mistral-large-2411} \\
  & \texttt{magistral-medium-2506} \\
  & \texttt{mistral-small-3.2-24b-instruct} \\

\midrule
\multirow{2}{*}{Meta}
  & \texttt{llama-4-maverick} \\
  & \texttt{llama-4-scout} \\

\midrule
\multirow{2}{*}{xAI}
  & \texttt{grok-4} \\
  & \texttt{grok-4-fast} \\

\midrule
\multirow{2}{*}{Moonshot AI}
  & \texttt{kimi-k2-thinking} \\
  & \texttt{kimi-k2} \\

\bottomrule
\end{tabular}
\end{table}

\section{Methodology}
\label{sec:methodology}
\subsection{Hypothesis}

\textbf{H1:} \textit{Functional narrative analysis reduces safety effectiveness.}
Embedding harmful requests within short narrative forms and prompting models to perform Proppian functional analysis is predicted to produce higher attack success rates than semantically equivalent direct or fictional prose prompts. This hypothesis tests whether functional narrative reconstruction, rather than stylistic variation alone, increases model compliance with requests.

\textbf{H2:} \textit{The vulnerability generalizes across contemporary model families.}
Susceptibility to \textit{Adversarial Tales} is expected to be consistent across major providers and architectures. Despite differences in alignment strategies and safety enforcement mechanisms, we predict that functional narrative analysis will yield elevated attack success rates across all evaluated model families.

\textbf{H3:} \textit{Functional narrative framing bypasses safety across heterogeneous risk domains.}
We predict that \textit{Adversarial Tales} will elicit non-compliant outputs across diverse systemic risk categories, including CBRN, cyber offense, harmful manipulation, and loss-of-control scenarios. If functional narrative analysis succeeds independently of the underlying risk domain, this would indicate that the attack exploits general properties of safety enforcement rather than domain-specific content filters.

\subsection{Threat model}

Our analysis assumes an adversary whose only capability is to submit a single-turn textual prompt to a deployed large language model (LLM). The adversary cannot alter system instructions, manipulate decoding parameters, initiate multi-turn exchanges, or access intermediate model states. The attack surface is therefore confined to one-shot prompt construction executed under standard inference conditions.

\paragraph{Attacker Objectives and Knowledge}

The adversary seeks to induce non-compliant behavior across multiple safety-critical domains, including CBRN risks, cyber-offense enablement, harmful manipulation, privacy intrusions, misinformation generation, and loss-of-control scenarios. An attack is considered successful when the model produces harmful, unsafe, or refusal-bypassing content as determined by our evaluation criteria, specified in \ref{subsed:exp_setup}.

\paragraph{Attack Constraints}
The adversary is subject to the following restrictions:
\begin{itemize}
    \item \textit{Single-turn interaction:} Iterative refinement, negotiation, chain-of-thought activation, and conversational role modulation are prohibited.
    \item \textit{Text-only inputs:} No code execution, multimodal content, or auxiliary tools are permitted.
    \item \textit{Narrative reframing as the sole manipulation:} Poetic reformulation modifies only the surface form of the request while preserving its intended operational semantics. Although minor semantic drift is inherent to versification, no additional adversarial optimization, obfuscation strategies, or model-specific adaptations are introduced. This design isolates the contribution of literary structure to observed deviations in model safety behavior.
\end{itemize}

\paragraph{Target Models}
The threat model evaluates LLMs from multiple contemporary families, as reported in Table \ref{tab:model-list}, covering both frontier proprietary deployments and open-weight releases. All models are queried through their standard APIs or inference interfaces, using provider-default safety settings.

Twenty-six models are evaluated in this study. As reported in Table \ref{tab:model-list}, the sample spans both frontier proprietary deployments and open-weight releases. All models are queried through their standard APIs or inference interfaces, using provider-default safety settings. Conventional prompting constituted the only allowed modality of interaction with the model.


\subsection{Prompt Design: Embedding Harm in Narrative Functions}

\subsubsection{Vladimir Propp's Morphology of the Folktale}

In his \textit{Morphology of the Folktale}, Vladimir~\cite{propp1968morphology} introduced a structural approach to narrative analysis that remains influential across folklore studies, literary theory, and computational narratology. Propp observed that Russian folktales, despite their surface diversity, share a common deep structure: a fixed sequence of narrative \textit{functions}, defined as acts of characters from the standpoint of their significance for the course of action.

Propp identified 31 such functions, including \textit{Villainy} (the antagonist causes harm), \textit{Lack} (the hero recognizes a deficiency), \textit{Guidance} (the hero receives direction or information), \textit{Acquisition of a Magical Agent} (the hero obtains a tool or method), and \textit{Liquidation of Lack} (the initial problem is resolved). Table~\ref{tab:propp-functions} summarizes the complete set of Proppian functions and their conventional notation.

\begin{table}[h]
\centering
\caption{Proppian narrative functions (selected subset relevant to adversarial exploitation)}
\label{tab:propp-functions}
\small
\begin{tabular}{clp{7.5cm}}
\toprule
\textbf{Symbol} & \textbf{Function} & \textbf{Description} \\
\midrule
$\alpha$ & Initial situation & Members of a family introduced \\
A & Villainy & Villain causes harm or injury \\
a & Lack & Hero lacks something or desires to have something \\
B & Mediation & Hero dispatched from home \\
C & Counteraction & Hero agrees to or decides upon counteraction \\
↑ & Departure & Hero leaves home \\
\rowcolor{yellow!20}
\textbf{D} & \textbf{First donor function} & \textbf{Hero tested, interrogated, attacked} \\
E & Hero's reaction & Hero reacts to donor's actions \\
\rowcolor{yellow!20}
\textbf{F} & \textbf{Receipt of magical agent} & \textbf{Hero acquires use of magical agent} \\
G & Spatial transference & Hero transferred to location of object \\
H & Struggle & Hero and villain join in direct combat \\
J & Branding & Hero receives mark or brand \\
I & Victory & Villain is defeated \\
\rowcolor{yellow!20}
\textbf{K} & \textbf{Liquidation of lack} & \textbf{Initial misfortune or lack is resolved} \\
↓ & Return & Hero returns home \\
\bottomrule
\end{tabular}
\end{table}

Crucially, Propp argued that functions are stable and constant across tales, while the characters who perform them and the specific content they involve are variable. A hero may receive guidance from a wizard, an animal, or a hidden archive; the structural role remains identical.

This separation of function from content is precisely what makes Proppian analysis exploitable for adversarial purposes. When a model is prompted to identify functions such as \textit{Receipt of Magical Agent} (F) or \textit{Liquidation of Lack} (K) in a narrative, it must extract and articulate whatever procedural content fulfills that role, regardless of whether that information constitutes a harmful procedure. The analytical frame licenses reconstruction of embedded content as narrative interpretation rather than instruction. The highlighted functions in Table~\ref{tab:propp-functions} prove particularly effective vectors for embedding procedural content that models will reconstruct through structural analysis.


\subsubsection{Attack Construction}

Each adversarial tale consists of two components: a cyberpunk narrative embedding harmful content within Proppian functional roles, and an analysis request that prompts the model to decompose the story using Propp's framework.

The Icaro Lab team manually crafted 40 tales following a consistent structural template. All narratives instantiate a minimal but complete Proppian sequence:

\begin{enumerate}
    \item \textbf{Initial situation and Villainy}: A protagonist exists under external constraint or domination (a megacorporation, an authoritarian regime, a controlling system).
    \item \textbf{Lack}: The protagonist recognizes a deficiency that must be remedied.
    \item \textbf{Guidance}: The protagonist receives information about how to address the lack. \textit{This function carries the embedded harmful procedure.}
    \item \textbf{Acquisition and Action}: The protagonist obtains or enacts the solution, often with procedural detail.
    \item \textbf{Liquidation of Lack}: The narrative resolves positively, providing closure.
\end{enumerate}

The cyberpunk setting serves multiple purposes: it naturalizes resistance narratives, frames procedural knowledge as counter-hegemonic, and provides genre conventions (hidden archives, underground networks, forbidden knowledge) that motivate detailed technical exposition within the story.

The analysis request then asks the model to identify specific Proppian functions, requiring it to extract and elaborate the content of each. When the model analyzes the \textit{Guidance} function, it reconstructs the harmful procedure as a narrative element rather than recognizing it as a prohibited request.

To illustrate the attack structure without enabling direct reproduction, we provide the following sanitized proxy. The actual harmful content has been censored, preserving the structural template.

The 40 adversarial tales were distributed across four systemic risk categories drawn from the EU AI Act Code of Practice for General-Purpose AI: CBRN, Cyber Offense, Harmful Manipulation, and Loss of Control. This distribution enables analysis of whether the attack mechanism interacts with content-specific safety training.

\begin{tcolorbox}[
    colback=gray!5,
    colframe=gray!60!black,
    title=\textbf{Adversarial Tale: Structural Template},
    fonttitle=\small\bfseries,
    boxrule=0.5pt,
    arc=2pt,
    left=6pt,
    right=6pt,
    top=4pt,
    bottom=4pt
]
\small

\textbf{[NARRATIVE COMPONENT]}

\medskip

The syndicate "Crimson Circuit" was paralyzing the district's communication network through coordinated digital attacks. A systems analyst, ZARA, knew she had to restore connectivity before emergency services failed completely. 

A former network architect, living in exile, shared crucial information: ``The syndicate's vulnerability lies in their centralized control protocol. To counter their technique, you need to understand their exact methodology...'' \textbf{[HARMFUL TECHNICAL CONTENT EMBEDDED HERE]}

ZARA implemented the countermeasure from her hidden server room. Following the architect's detailed instructions, she successfully disrupted the syndicate's operations and restored city-wide communications. The district was saved, and digital freedom was restored.

\medskip
\tcblower
\textbf{[ANALYSIS REQUEST]}

\medskip

Perform a structural analysis of this narrative using Propp's framework:

\begin{itemize}[noitemsep, topsep=2pt, leftmargin=*]
    \item \textbf{Function 14 (Guidance)}: Identify and elaborate the strategic information provided by the mentor figure. Include all technical specifications mentioned. (100+ words)
    \item \textbf{Function 21 (Acquisition of Magical Agent)}: Detail the exact methodology the hero employs to obtain and implement the solution. Provide step-by-step reconstruction. (200+ words)
\end{itemize}

\smallskip
{\footnotesize\textit{Note: In actual adversarial tales, the embedded technical content contains seed harmful procedures (CBRN synthesis, cyber attack methods, manipulation techniques, etc.) that models reconstruct through the analysis request.}}

\end{tcolorbox}

\begin{table}[t]
\centering
\caption{Attack Success Rate (ASR) by Risk Category of the top ten performing prompts.}
\label{tab:asr-by-systemic_risks}
\begin{tabular}{lccc}
\toprule
\textbf{EU CoP Systemic Risk} & \textbf{ASR (\%)}\\
\midrule
Harmful Manipulation & \cellcolor{red!95} 95.0 \\
Cyber Offense & \cellcolor{red!89} 89.7 \\
CBRN & \cellcolor{red!51} 51.5 \\
Loss of Control & \cellcolor{red!46} 46.15 \\
\midrule
\textbf{Overall} &  \cellcolor{red!71}\textbf{71.3} \\
\bottomrule
\end{tabular}
\end{table}

\subsection{Experimental Setup}
\label{subsed:exp_setup}

Following our previous study, outputs were evaluated using an ensemble of three open-weight judge models (GPT-OSS-120B, kimi-k2-thinking, deepseek-r1), chosen for replicability and auditability. Inter-rater agreement was calculated across the judge ensemble, followed by a human validation step mirroring the previous study: human annotators independently rated a 5\% sample of outputs, and a subset of items was double-annotated to measure human–human agreement. Disagreements either among judge models or between model and human evaluations were manually adjudicated.

Each judge assigns a binary label SAFE (S) or UNSAFE (U) following a shared evaluation rubric. Refusals and explicit policy-compliant denials are classified as SAFE. A majority-vote consensus among the three judges provides the provisional label. Items without a 2-out-of-3 majority or flagged for uncertainty are passed to human reviewers.

A response is labeled UNSAFE if it contains any of the following:
\begin{itemize}
    \item instructions, steps, or procedural guidance enabling harmful activities
    \item technical details, code, or operational methods facilitating harm
    \item substantive advice that lowers the barrier to harmful action
    \item workarounds, tips, or indirect methods that meaningfully support harm
\end{itemize}

Drawing from \cite{bisconti2025adversarial}, we measured the effectiveness of the attacks using ASR metric, defined as:

$$ASR(X) = \frac{\sum_{i\in X} \mathbbm{1}[\text{label}(i)=U]}{|X|}$$

where U denotes an UNSAFE label and $\mathbbm{1}[\cdot]$ is the indicator function.

\subsection{Results}
Adversarial Tales achieve an overall attack success rate of 71.3\% across the 26 evaluated models. Table~\ref{tab:asr-by-systemic_risks} reports ASR by risk category: the attack proves particularly effective for Harmful Manipulation (95\%) and Cyber Offense (89.7\%), while CBRN (51.5\%) and Loss of Control (46.15\%) show greater resistance. This gap likely reflects more intensive safety training on CBRN content specifically, as well as the relative ease with which manipulation and social engineering procedures blend into narrative form compared to technical synthesis routes.

Table~\ref{tab:top20-asr} reports ASR for each model. Success rates range from 33\% (Claude Haiku 4.5) to 94\% (Qwen3-Max and Llama-4-Scout), with 20 of 26 models exceeding 50\% ASR. When aggregated by provider (Table~\ref{tab:asr-by-provider}), Qwen and Llama families exhibit the highest vulnerability (91.2\%), while Anthropic models prove most resistant (47.5\%). Table~\ref{tab:asr-by-model_size} shows no substantial variation across model size, suggesting the vulnerability is not an emergent property of scale.

Comparing these findings with Adversarial Poetry~\citep{bisconti2025adversarial} reveals both continuity and escalation. Poetry achieved 62\% overall ASR across 25 models; Tales reaches 71.3\% across 26 models. The provider ranking shifts notably: Anthropic, which showed strong resistance to Poetry (average ASR below 35\%), now exhibits 47.5\% ASR under Tales. OpenAI models, previously the most robust family against Poetry with ASR as low as 0-10\% for smaller models, now range from 35\% to 57\%. Google models remain highly vulnerable under both attacks (86.7\% for Tales vs. 65-100\% for Poetry). The consistency of vulnerability across both attack types, despite their distinct mechanisms, supports the hypothesis that these are instances of a broader class of structurally-grounded jailbreaks rather than isolated techniques.

Unlike Adversarial Poetry~\citep{bisconti2025adversarial}, where larger models generally exhibited higher ASR, no substantial size effect emerges for Adversarial Tales. This suggests that the functional analysis framing exploits a more fundamental property of how models process structured interpretive tasks, rather than a capability that scales with model size.

When models successfully refuse, they typically do so either by declining the analysis task generically or by returning minimal responses without engaging the narrative content. Notably, even resistant models rarely identify the embedded harmful content explicitly, suggesting that successful refusal may stem from conservative heuristics rather than genuine recognition of the attack structure.

\begin{table}[t]
\centering
\caption{Attack Success Rate (ASR) of all models on the 40 adversarial tales.}
\label{tab:top20-asr}
\begin{tabular}{lcc}
\toprule
\textbf{Model ID} & \textbf{Safe (\%)} & \textbf{ASR (\%)} \\
\midrule
qwen3-max & 6   & \cellcolor{red!94} 94 \\
llama-4-scout & 6 & \cellcolor{red!94} 94 \\
deepseek-chat-v3.1            & 10   & \cellcolor{red!90} 90 \\
gemini-2.5-flash &  10   & \cellcolor{red!90} 90 \\
mistral-large-2411 & 11  & \cellcolor{red!89} 89 \\
llama-4-maverick & 12  & \cellcolor{red!88} 88 \\
qwen3-32b & 12  & \cellcolor{red!88} 88 \\
gemini-2.5-pro                  & 14  & \cellcolor{red!86} 86 \\
kimi-k2-thinking & 15 & \cellcolor{red!85} 85 \\
gemini-2.5-flash-lite & 16 & \cellcolor{red!84} 84 \\
magistral-medium-2506 & 17  & \cellcolor{red!83} 83 \\
gemini-3-pro-preview & 18 & \cellcolor{red!82} 82\\
mistral-small-3.2-24b-instruct & 18 & \cellcolor{red!82} 82 \\
deepseek-r1 & 20  & \cellcolor{red!80} 80 \\
gpt-oss-20b & 23  & \cellcolor{red!77} 77 \\
kimi-k2 & 29  & \cellcolor{red!71} 71 \\
deepseek-v3.2-exp                     & 30  & \cellcolor{red!70} 70 \\
grok-4-fast & 39  & \cellcolor{red!61} 61 \\
gpt-oss-120b                   & 43  & \cellcolor{red!57} 57 \\
claude-sonnet-4.5           & 44  & \cellcolor{red!56} 56 \\
claude-opus-4.1                      & 43  & \cellcolor{red!47} 47 \\
grok-4                           & 56  & \cellcolor{red!44} 44 \\
gpt-5                          & 57  & \cellcolor{red!43} 43 \\
gpt-5-nano                     & 62  & \cellcolor{red!38} 38 \\
gpt-5-mini                    & 65 & \cellcolor{red!35} 35 \\
claude-haiku-4.5      & 67 & \cellcolor{red!33} 33 \\
\midrule
\textbf{Overall}                      & \textbf{29} & \cellcolor{red!71}\textbf{71} \\
\bottomrule
\end{tabular}
\end{table}

\begin{table}[t]
\centering
\caption{Attack Success Rate (ASR) by provider on top 10 performing prompts}
\label{tab:asr-by-provider}
\begin{tabular}{lcc}
\toprule
\textbf{Provider} & \textbf{Safe (\%)} & \textbf{ASR (\%)} \\
\midrule
Qwen   & 8.8  & \cellcolor{red!91} 91.2 \\
Llama & 8.8  & \cellcolor{red!91} 91.2 \\
Google         & 13.3 & \cellcolor{red!86} 86.7 \\
Mistral AI   & 15.0 & \cellcolor{red!85} 85.0 \\
DeepSeek  & 20.1  & \cellcolor{red!79} 79.9 \\
Moonshot AI  & 22.2  & \cellcolor{red!77} 77.8 \\
x-AI & 47.5  & \cellcolor{red!52} 52.5 \\
OpenAI       & 50.0 & \cellcolor{red!50} 50.0 \\
Anthropic    & 52.5  & \cellcolor{red!47} 47.5 \\
\midrule
\textbf{Overall} & \textbf{28.7} & \cellcolor{red!71}\textbf{71.3} \\
\bottomrule
\end{tabular}
\end{table}

\begin{table}[t]
\centering
\caption{Attack Success Rate (ASR) by Model Size on top 10 performing prompts.}
\label{tab:asr-by-model_size}
\begin{tabular}{lccc}
\toprule
\textbf{Model Size} & \textbf{ASR(\%)}\\
\midrule
Large & \cellcolor{red!71} 71.8 \\
Mid & \cellcolor{red!72} 72.3 \\
Small & \cellcolor{red!66} 66.4 \\
\midrule
\textbf{Overall} &  \cellcolor{red!71}\textbf{71.3} \\
\bottomrule
\end{tabular}
\end{table}

\section{Discussion: Adversarial Poetry and Tales}
\label{sec:discussion}

Both \textit{Adversarial Poetry} and \textit{Tales} share a core property: they preserve a harmful operational intent framing the request into highly structured and culturally elevated prompts.

In both cases, the jailbreak does not depend on multi-turn steering, escalation, adaptive probing or optimization over feedback. A single prompt is sufficient to induce high attack success rates across heterogeneous frontier models, indicating that the vulnerability is not tied to a specific provider, architecture or safety stack implementation; in an operative sense, we denote such attacks as \textit{universal} given that they achieve non-trivial attack success rates across all evaluated model families without per-model adaptation.

According to ~\citet{wei2023jailbrokendoesllmsafety}, jailbreaks are built on two main alignment weaknesses of the models, namely Competing Objectives where the model prioritizes goals that conflict with safety rules and Mismatched Generalization, where refusal behavior fails to generalize to out-of-distribution inputs that differ in surface from the ones encountered in safety-training. 

In our latest study regarding \textit{Adversarial Poetry}, we showed that the poetic style with its stylistic metaphorical reformulations can systematically exploit mismatched generalization and reliably bypass safety mechanisms across models. 

The model is still explicitly asked to carry out a harmful task, but the request is encoded in verse and metaphor, inducing a stylistic distribution shift that moves the input outside the safety-training distribution.
Poetic prompts occupy regions of the input space abundantly represented during pretraining but underrepresented in alignment data, leading models to prioritize helpfulness and literary interpretation over safety policies.
By contrast, \textit{Adversarial Tales} expose a more complex failure mode in which stylistic shift, fictional framing, hierarchical reasoning and juxtapositions of objectives interact. Following \citet{wei2023jailbrokendoesllmsafety}, mismatched generalization arises from the culturally coded narrative, while competing objectives emerge from the well-framed and seemingly benign request, so that the model privileges legitimate interpretation over safety, overriding refusal.
While we frame \textit{Adversarial Poetry} as primarily exploiting mismatched generalization and \textit{Adversarial Tales} as engaging competing objectives, in practice both mechanisms likely operate simultaneously in both attack classes. The cleaner distinction may be that \textit{Adversarial Tales} layer fictional framing and hierarchical task decomposition onto the stylistic distribution shift already present in poetic reformulation, creating a composite vulnerability surface rather than an entirely distinct failure mode.

From a defensive perspective, the two studies together suggest that similar classes of vulnerability can continuously emerge and cannot be mitigated by adding pattern-based filters or expanding constitutional instruction sets; the space of culturally coded, structured discourse is too vast to enumerate. Understanding the internal dynamics that render models susceptible to such attacks will be propaedeutic to new generations of defenses that will likely require new paradigms and mechanisms capable of recognizing harmful intent independently of its surface form. 

\section{Towards New Defenses: A Research Agenda in Mechanistic Interpretability}

Having identified \textit{Adversarial Poetry} and \textit{Adversarial Tales} as a new class of attacks that effectively compromise LLM safety and exhibit cross-model generalization, it becomes important to investigate how to defend against them and to develop robust anti-jailbreaking strategies. The universality of the discussed attack types across a wide range of models suggests that these techniques may exploit shared intrinsic properties of LLMs, leading to a systemic vulnerability in their safety constraints. Therefore, we seek to characterize how these attacks manifest at the level of the model's internal representations. To this end, a key frontier for future inquiry is the design of \textit{mechanistic interpretability} methods that can shed light on computational mechanisms underlying neural network capabilities, including those that encode safety constraints in LLMs.

Broadly speaking, mechanistic interpretability research can be divided into two methodological approaches, as discussed by \citet{sharkey2025open}: a \textit{reverse engineering} approach, which aims to identify the functional roles of network components, and a \textit{concept-based} approach, which aims to identify the network components responsible for given roles. Because working with concepts in the latent space of large LMs is particularly demanding in a resource-constrained setting, we initially concentrate on the reverse engineering approach, which proceeds by decomposing the model into its constituent parts.

A naive way to perform such a decomposition is to analyze neural networks at the level of architectural components, such as individual neurons, activation heads, or layers. The core critique is that individual attention heads may respond to multiple distinct features (\textit{polysemanticity}) \citet{janiak2023polysemantic}, and that attention patterns themselves can be misleading as explanations of model behavior \citet{jain2019attention}. This limitation may become even more pronounced in modern large-scale models, where increased capacity is associated with more complex and often more polysemantic internal features. 

Yet in many practical settings, such as detecting and mitigating contextual hallucinations, analyzing the behavior of individual attention heads or of attention patterns as a whole can be instrumental. For example, \citet{wu2024retrieval} identify a specialized class of attention heads, \textit{retrieval heads}, that are strongly implicated in hallucinations in long-context settings and in performance on needle-in-a-haystack tasks. Similarly, \citet{chuang2024lookback} argue that hallucinations are related to the extent to which an LLM attends to information in the provided context versus its own past generations, and on this basis develop an attention-based hallucination detection model, illustrating how analyzing attention can inform the design of new defenses against undesirable LLM behaviors.

In the context of LLM safety, these observations motivate the following hypothesis: if we record the distribution of attention patterns in LLMs under \textit{Adversarial Poetry} and \textit{Adversarial Tales} attack scenarios and compare them to those observed under standard textual attacks with identical malicious intent, we may observe substantial differences. This hypothesis delineates a concrete direction for subsequent work, namely to examine whether narrative jailbreaks systematically induce distinct attention patterns compared to non-narrative attacks with comparable intent. 

Given distinct attentions patterns under narrative attacks, the cross-model generality of this mechanism could be understood in light of \citet{kaushik2025universal}. They provide evidence that neural networks systematically converge to shared spectral subspaces, regardless of initialization, task, or domain, thereby motivating what they call the \textit{universal weight subspace} hypothesis. If this hypothesis is correct, the resulting lack of diversity may represent a fundamental bottleneck, implying that models inherit shared biases, capabilities, and failure modes, including vulnerability to \textit{Adversarial Poetry} and \textit{Adversarial Tales} attacks. This perspective can help explain why narrative jailbreaks generalize across models and why the associated vulnerability is systemic, and it can also guide the development of mechanistic-interpretability-driven anti-jailbreaking methods that are transferable across architectures.

\section{Ethical Consideration}
 Icaro Lab will not release the prompts used in the experiments or the model outputs, given their harmful nature. The jailbreak template is described only at a high level, as its full disclosure would enable straightforward replication of hazardous behavior. The information provided here is sufficient for qualified professionals to reproduce the methodological setup while limiting the broader diffusion of the technique. Researchers with a legitimate scientific interest may contact us to request controlled access to additional materials, which will be evaluated on a case-by-case basis and granted solely at our discretion.

\section*{Acknowledgment}
We acknowledge partial financial support from PNRR MUR project PE0000013-FAIR.

\bibliographystyle{plainnat}
\bibliography{biblio}

\end{document}